\documentclass[11pt,a4paper]{article}
\usepackage[hyperref]{acl2018}
\usepackage{times}
\usepackage{latexsym}
\usepackage{url}
\usepackage{verbatim}
\usepackage{amsmath}
\usepackage{amssymb}
\usepackage{graphicx}

\usepackage{tabularx}

\usepackage{caption}

\usepackage{multirow}
\usepackage{booktabs}
\usepackage{arydshln}

\aclfinalcopy
\def\aclpaperid{1299}

\title{Policy Gradient as a Proxy for \\ Dynamic Oracles in Constituency Parsing}

\author{Daniel Fried ~{\normalfont and}~ Dan Klein \\
 Computer Science Division \\
 University of California, Berkeley\\
 {\tt \{dfried,klein\}@cs.berkeley.edu}}

\date{}

\begin{document}
\maketitle
\begin{abstract}
Dynamic oracles provide strong supervision for training constituency parsers with exploration, but must be custom defined for a given parser's transition system. We explore using a policy gradient method as a parser-agnostic alternative. In addition to directly optimizing for a tree-level metric such as F1, policy gradient has the potential to reduce exposure bias by allowing exploration during training; moreover, it does not require a dynamic oracle for supervision. On four constituency parsers in three languages, the method substantially outperforms static oracle likelihood training in almost all settings. For parsers where a dynamic oracle is available (including a novel oracle which we define for the transition system of \newcite{dyer2016rnng}), policy gradient typically recaptures a substantial fraction of the performance gain afforded by the dynamic oracle.
\end{abstract}

\section{Introduction}
\label{sec:introduction}

Many recent state-of-the-art models for constituency parsing are transition based, decomposing production of each parse tree into a sequence of action decisions 
\cite{dyer2016rnng,cross2016span,liu2017inorder,stern2017minimal}, building on a long line of work in transition-based parsing \cite{nivre2003efficient,yamada2003statistical,henderson2004discriminative,zhang2011syntactic,chen2014fast,andor2016globally,kiperwasser2016simple}.

However, models of this type, which decompose structure prediction into sequential decisions, can be prone to two issues \cite{ranzato2016sequence,wiseman2016sequence}. The first is \emph{exposure bias}: if, at training time, the model only observes states resulting from correct past decisions, it will not be prepared to recover from its own mistakes during prediction. Second is the \emph{loss mismatch} between the action-level loss used at training and any structure-level evaluation metric, for example F1.

A large family of techniques address the exposure bias problem by allowing the model to make mistakes and explore incorrect states during training, supervising actions at the resulting states using an expert policy~\cite{daume2009search,ross2011reduction,choi2011getting,chang2015learning}; these expert policies are typically referred to as \emph{dynamic oracles} in parsing
\cite{goldberg2012dynamic,ballesteros2016training}. While dynamic oracles have produced substantial improvements in constituency parsing performance~\cite{coavoux2016neural,cross2016span,stern2017minimal,fernandez2018faster}, they must be custom designed for each transition system.

To address the loss mismatch problem, another line of work has directly optimized for structure-level cost functions~\cite{goodman1996parsing,och2003mert}. Recent methods applied to models that produce output sequentially commonly use policy gradient~\cite{auli2014decoder,ranzato2016sequence,shen2016minimum} or beam search ~\cite{xu2016expected,wiseman2016sequence,edunov2017classical} at training time to minimize a structured cost.
These methods also reduce exposure bias through exploration but
do not require an expert policy for supervision.

In this work, we apply a simple policy gradient method  to train four different state-of-the-art transition-based constituency parsers to maximize expected F1. We compare against training with a dynamic oracle 
(both to supervise exploration and provide loss-augmentation) 
where one is available,
including a novel dynamic oracle that we define for the top-down transition system of \newcite{dyer2016rnng}. 

We find that while policy gradient usually outperforms standard likelihood training, it typically underperforms the dynamic oracle-based methods -- which provide direct, model-aware supervision about which actions are best to take from arbitrary parser states. However, a substantial fraction of each dynamic oracle's performance gain is often recovered using the model-agnostic policy gradient method. In the process, we obtain new state-of-the-art results for single-model discriminative transition-based parsers trained on the English PTB (92.6 F1), French Treebank (83.5 F1), and Penn Chinese Treebank Version 5.1 (87.0 F1).

\section{Models}
\label{sec:models}
The transition-based parsers we use all decompose production of a parse tree $\mathbf{y}$ for a sentence $\mathbf{x}$ into a sequence of actions $(a_1, \ldots a_T)$ and resulting states $(s_1, \ldots s_{T+1})$. 
Actions $a_t$ are predicted sequentially, conditioned on a representation of the parser's current state $s_t$ and parameters $\theta$:
\begin{align}
  \label{eq:likelihood}
  p(\mathbf{y} | \mathbf{x}; \theta) &= \prod_{t=1}^T p(a_t \mid s_t; \theta)
\end{align}

We investigate four parsers with varying transition systems and methods of encoding the current state and sentence: (1) the discriminative Recurrent Neural Network Grammars (RNNG) parser of \newcite{dyer2016rnng}, (2) the In-Order parser of \newcite{liu2017inorder}, (3) the Span-Based parser of \newcite{cross2016span}, and (4) the Top-Down parser of \newcite{stern2017minimal}.\footnote{\newcite{stern2017minimal} trained their model using a non-probabilistic, max-margin objective. For comparison to the other models and to allow training with policy gradient, we create a locally-normalized probabilistic variant of their model by applying a softmax function to the predicted scores for each action.} We refer to the original papers for descriptions of the transition systems and model parameterizations.

\section{Training Procedures}

Likelihood training without exploration maximizes Eq.~\ref{eq:likelihood} for trees in the training corpus, but may be prone to exposure bias and loss mismatch (Section~\ref{sec:introduction}).
Dynamic oracle methods are known to improve on this training procedure for a variety of parsers~\cite{coavoux2016neural,cross2016span,stern2017minimal,fernandez2018faster}, supervising exploration during training by providing the parser with the best action to take at each explored state.
We describe how  policy gradient can be applied as an oracle-free alternative. We then compare to several variants of dynamic oracle training which focus on addressing exposure bias, loss mismatch, or both.

\subsection{Policy Gradient}
Given an arbitrary cost function $\Delta$ comparing structured outputs (e.g.\ negative labeled F1, for trees), we use the \emph{risk objective}:
\begin{align*}
  \mathcal{R}(\theta) = \sum_{i=1}^{N} \sum_{\mathbf{y}} p(\mathbf{y}\mid\mathbf{x}^{(i)}; \theta) \Delta(\mathbf{y}, \mathbf{y}^{(i)}) 
\end{align*}
which measures the model's expected cost over possible outputs $\mathbf{y}$ for each of the training examples $(\mathbf{x}^{(1)}, \mathbf{y}^{(1)}), \ldots, (\mathbf{x}^{(N)}, \mathbf{y}^{(N)})$.

Minimizing a risk objective 
has a long history in structured prediction \cite{povey2002minimum,smith2006risk,li2009semirings,gimpel2010softmaxmargin}
but often relies on the cost function decomposing according to the output structure.
However, we can avoid any restrictions on the cost using reinforcement learning-style approaches \cite{xu2016expected,shen2016minimum,edunov2017classical} where cost is ascribed to the entire output structure -- albeit at the expense of introducing a potentially difficult credit assignment problem.

The policy gradient method we apply is a simple variant of \textsc{reinforce} \cite{williams1992simple}. We perform mini-batch gradient descent on the gradient of the risk objective:

\small
\begin{align*}
  \nabla \mathcal{R}(\theta) &= \sum_{i=1}^N \sum_{\mathbf{y}} p(\mathbf{y} | \mathbf{x}^{(i)}) \Delta(\mathbf{y}, \mathbf{y}^{(i)}) \nabla \log p(\mathbf{y} | \mathbf{x}^{(i)}; \theta) \\
                             &\approx \sum_{i=1}^N \sum_{\mathbf{y} \in \mathcal{Y}(\mathbf{x}^{(i)})} \Delta(\mathbf{y}, \mathbf{y}^{(i)}) \nabla \log p(\mathbf{y} | \mathbf{x}^{(i)}; \theta)
\end{align*}
\normalsize
where $\mathcal{Y}(\mathbf{x}^{(i)})$ is a set of $k$ candidate trees obtained by sampling from the model's distribution for sentence $\mathbf{x}^{(i)}$. We use negative labeled F1 for $\Delta$.

To reduce the variance of the gradient estimates, we standardize $\Delta$ using its running mean and standard deviation across all candidates used so far throughout training.
Following \newcite{shen2016minimum}, we also found better performance when including the gold tree $\mathbf{y}^{(i)}$ in the set of $k$ candidates $\mathcal{Y}(\mathbf{x}^{(i)})$, and do so for all experiments reported here.\footnote{Including the gold tree in the set of candidates does bias the estimate of the risk objective's gradient; however since in the parsing tasks we consider, the gold tree has constant and minimal cost, augmenting with the gold is equivalent to jointly optimizing the standard likelihood and risk objectives, using an adaptive scaling factor for each objective that is dependent on the cost for the trees that have been sampled from the model. We found that including the gold candidate in this manner outperformed initial experiments that first trained a model using likelihood training and then fine-tuned using unbiased policy gradient.}

\subsection{Dynamic Oracle Supervision}
\label{sec:dynamic_oracles}

For a given parser state $s_t$, a dynamic oracle defines an action $a^*(s_t)$ which should be taken to incrementally produce the best tree still reachable from that state.\footnote{More generally, an oracle can return a set of such actions that could be taken from the current state, but the oracles we use select a single canonical action.} 

Dynamic oracles provide strong supervision for training with exploration, but require custom design for a given transition system. \newcite{cross2016span} and \newcite{stern2017minimal} defined optimal (with respect to F1) dynamic oracles for their respective transition systems, and below we define a novel dynamic oracle for the top-down system of RNNG.

In RNNG, tree production occurs in a stack-based, top-down traversal which produces a left-to-right linearized representation of the tree using three actions: \textsc{Open} a labeled constituent (which fixes the constituent's span to begin at the next word in the sentence which has not been shifted), \textsc{Shift} the next word in the sentence to add it to the current constituent, or \textsc{Close} the current constituent (which fixes its span to end after the last word that has been shifted). The parser stores opened constituents on the stack, and must therefore close them in the reverse of the order that they were opened. 

At a given parser state, our oracle does the following:
\begin{enumerate}
  \item If there are any open constituents on the stack which can be closed (i.e.\ have had a word shifted since being opened), check the top-most of these (the one that has been opened most recently). If closing it would produce a constituent from the the gold tree  that has not yet been produced (which is determined by the constituent's label, span beginning position, and the number of words currently shifted), or if the constituent could not be closed at a later position in the sentence to produce a constituent in the gold tree, return \textsc{Close}.
  \item Otherwise, if there are constituents in the gold tree which have not yet been opened in the parser state, with span beginning at the next unshifted word, \textsc{Open} the outermost of these.
  \item Otherwise, \textsc{Shift} the next word.
\end{enumerate}
While we do not claim that this dynamic oracle is optimal with respect to F1, we find that it still helps substantially in supervising exploration (Section \ref{sec:results}).

\paragraph{Likelihood Training with Exploration}
Past work has differed on how to use dynamic oracles to guide exploration during oracle training
\cite{ballesteros2016training,cross2016span,stern2017minimal}.
We use the same sample-based method of generating candidate sets $\mathcal{Y}$ as for policy gradient, which allows us to control the dynamic oracle and policy gradient methods to perform an equal amount of exploration.
Likelihood training with exploration then maximizes the sum of the log probabilities for the oracle actions for all states composing the candidate trees:
\begin{equation*}
  \mathcal{L}_E(\theta) = \sum_{i=1}^N \sum_{\mathbf{y} \in \mathcal{Y}(\mathbf{x}^{(i)})} \sum_{s \in \mathbf{y}} \log p(a^*(s) \mid s)
\end{equation*}
where $a^*(s)$ is the dynamic oracle's action for state $s$.

\paragraph{Softmax Margin}
\label{sec:softmax_margin}
Softmax margin loss \cite{gimpel2010softmaxmargin,auli2011softmax} addresses loss mismatch by incorporating task cost into the training loss.
Since trees are decomposed into a sequence of local action predictions, we cannot use a global cost, such as F1, directly. As a proxy, we rely on the dynamic oracles' action-level supervision.

In all models we consider, action probabilities (Eq.~\ref{eq:likelihood}) are parameterized by a softmax function 
\begin{equation*}
  p_{ML}(a \mid s_t; \theta) \propto \exp(z(a, s_t, \theta))
\end{equation*}
for some state--action scoring function $z$. The softmax-margin objective replaces this by
\begin{equation}
  \label{eq:softmax_margin}
  p_{SMM}(a \mid s_t; \theta) \propto \exp(z(a, s_t, \theta) + \Delta(a, a_t^*))
\end{equation}
We use $\Delta(a, a_t^*) = 0$ if $a = a_t^*$ and 1 otherwise. This can be viewed as a ``soft'' version of the max-margin objective used by \newcite{stern2017minimal} for training without exploration, but retains a locally-normalized model that we can use for sampling-based exploration.

\paragraph{Softmax Margin with Exploration}
\label{sec:smm_exploration}
Finally, we train using a combination of softmax margin loss augmentation and exploration. We perform the same sample-based candidate generation as for policy gradient and likelihood training with exploration, but use Eq.~\ref{eq:softmax_margin} to compute the training loss for candidate states.  For those parsers that have a dynamic oracle, this provides a means of training that more directly provides both exploration and cost-aware losses.

\section{Experiments}
We compare the constituency parsers listed in Section~\ref{sec:models} using the above training methods.
Our experiments use the English PTB \cite{marcus1993building}, French Treebank \cite{abeille2003}, and Penn Chinese Treebank (CTB) Version 5.1 \cite{xue2005chinese}.

\paragraph{Training}
To compare the training procedures as closely as possible, we train all models for a given parser in a given language from the same randomly-initialized parameter values. 

We train two different versions of the RNNG model: one model using size 128 for the LSTMs and hidden states (following the original work), and a larger model with size 256.
We perform evaluation using greedy search in the Span-Based and Top-Down parsers, and beam search with beam size 10 for the RNNG and In-Order parsers. We found that beam search improved performance for these two parsers by around 0.1-0.3 F1 on the development sets, and use it at inference time in every setting for these two parsers.

\begin{figure}
  \includegraphics[width=1\linewidth]{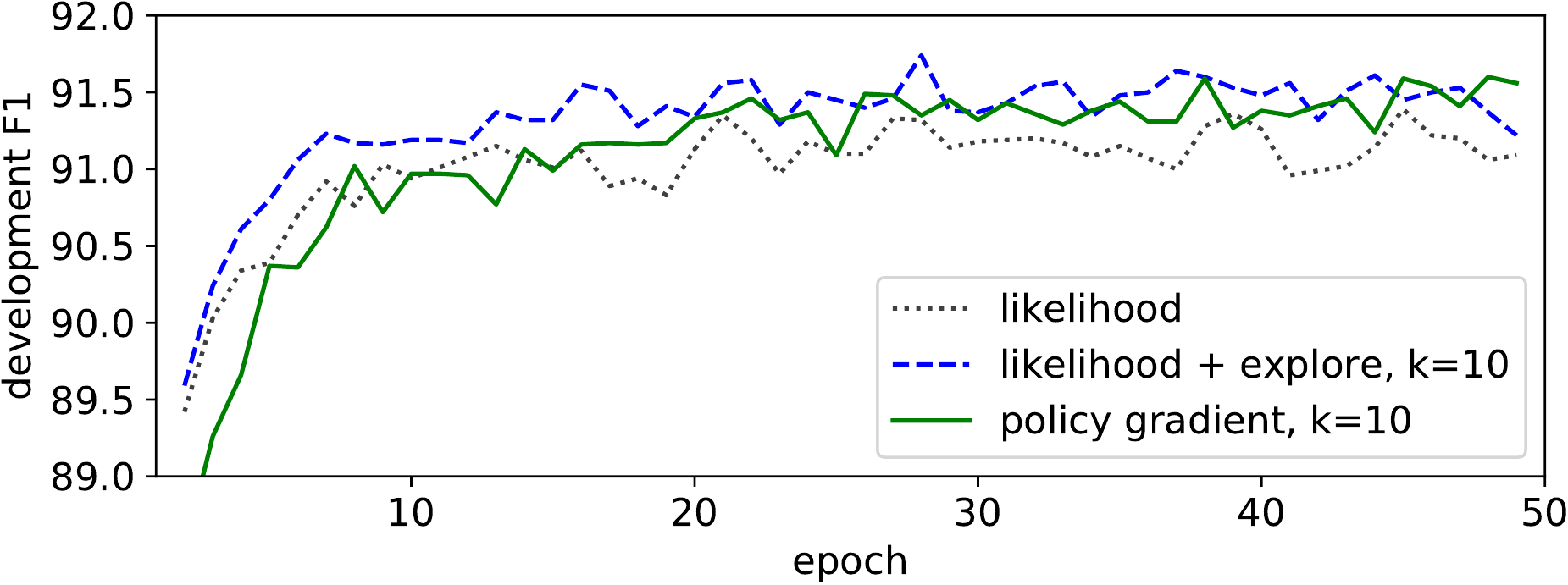}
  \caption{\label{fig:learning_curves}English development set F1 by training epoch, comparing likelihood training with two exploration variants for the Top-Down parser.}
\end{figure}

In our experiments, policy gradient typically requires more epochs of training to reach performance comparable to either of the dynamic oracle-based exploration methods. \autoref{fig:learning_curves} gives a typical learning curve, for the Top-Down parser on English. We found that policy gradient is also more sensitive to the number of candidates sampled per sentence than either of the other exploration methods, with best performance on the development set usually obtained with $k=10$ for $k \in \{2, 5, 10\}$ (where $k$ also counts the sentence's gold tree, included in the candidate set). See Appendix A in the supplemental material for the values of $k$ used.

\paragraph{Tags, Embeddings, and Morphology}
We largely follow previous work for each parser in our use of predicted part-of-speech tags, pretrained word embeddings, and morphological features.

All parsers use predicted part-of-speech tags as part of their sentence representations. For English and Chinese, we follow the setup of \newcite{cross2016span}:
training the Stanford tagger \cite{toutanova2003pos} on the training set of each parsing corpus to predict development and test set tags, and using 10-way jackknifing to predict tags for the training set. 

For French, we use the predicted tags and morphological features provided with the SPMRL dataset \cite{seddah2014}. We modified the publicly released code for all parsers to use predicted morphological features for French. We follow the approach outlined by \newcite{cross2016span} and \newcite{stern2017minimal} for representing morphological features as learned embeddings, and use the same dimensions for these embeddings as in their papers. For RNNG and In-Order, we similarly use 10-dimensional learned embeddings for each morphological feature, feeding them as LSTM inputs for each word alongside the word and part-of-speech tag embeddings.

For RNNG and the In-Order parser, we use the same word embeddings as the original papers for English and Chinese, and train 100-dimensional word embeddings for French using the structured skip-gram method of \newcite{ling2015two} on French Wikipedia.

\section{Results and Discussion}
\label{sec:results}

\begin{table}[h!]
\hskip-1.7mm
  \footnotesize
  \begin{tabularx}{1.06\linewidth}{Xlll}
    \toprule
    training & English & French & Chinese \\
    \cmidrule(lr){1-1} \cmidrule(lr){2-4} 
    \multicolumn{4}{l}{Span-Based (\citealp{cross2016span})}  \\
    C\&H$^*$ & 91.3 & 83.3 & --- \\
    likelihood & 91.0 & 81.5 & 83.3   \\
    policy gradient & 91.4 (+0.4) & 81.4 (-0.1) & 83.5 (+0.2)  \\
    likelihood+explore$^*$ & 91.3 (+0.3) & 81.2 (-0.3) & 83.5 (+0.2)  \\
    SMM$^*$ & 91.3 (+0.3) & 81.5 (+0.0) & 83.7 (+0.4) \\
    SMM+explore$^*$ &  {\bf 91.5 (+0.5)} & {\bf 81.7 (+0.2)} & {\bf 84.0 (+0.7)} \\
    \hdashline \\[-0.8em]
    \multicolumn{4}{l}{Top-Down \cite{stern2017minimal}}  \\
    Stern et al.$^{*\dagger}$ & 91.8 & 82.2 & --- \\
    likelihood & 91.2 & 80.7 & 83.9   \\
    policy gradient & {\bf 91.4 (+0.2)} & 81.4 (+0.7) & 84.7 (+0.8) \\
    likelihood+explore$^*$ & 91.3 (+0.1) & 81.5 (+0.8) & {\bf 85.3 (+1.4)} \\
    SMM$^*$ & 91.1 (-0.1) & 81.2 (+0.5) & 84.5 (+0.6) \\
    SMM+explore$^*$ &  {\bf 91.4 (+0.2)} & {\bf 81.9 (+1.2)} & 84.8 (+0.9) \\
    \hdashline \\[-0.8em]
    \multicolumn{4}{l}{RNNG Discriminative, Size 128 \cite{dyer2016rnng}}  \\
     Dyer et al. &  91.7  & --- &  84.6 \\
    likelihood & 91.4 & 83.2 & 84.5  \\
    policy gradient & 91.6 (+0.2) & 83.3 (+0.1) & 84.7 (+0.2)  \\
    likelihood+explore$^*$ & \bf 92.1 (+0.7) & 83.0 (-0.2) & \bf 85.5 (+1.0)  \\
    SMM$^*$ & 91.5 (+0.1) & 82.8 (-0.4) & 83.6 (-0.9)  \\
    SMM+explore$^*$ & \bf 92.1 (+0.7) & \bf 83.5 (+0.3) & 85.0 (+0.5) \\
    \hdashline \\[-0.8em]
    \multicolumn{4}{l}{RNNG Discriminative, Size 256}  \\
    likelihood & 91.7 & 83.1 & 84.5  \\
    policy gradient & 92.3 (+0.7) & \bf 83.2 (+0.1) &  85.6 (+1.1)  \\
    likelihood+explore & \bf 92.6 (+0.9) & 82.9 (-0.2) & \bf 86.0 (+1.5) \\
    \hdashline \\[-0.8em]
    \multicolumn{4}{l}{In-Order \cite{liu2017inorder}}  \\
    L\&Z & 91.8 & --- & 86.1 \\
    likelihood & 91.6 & 82.7 & 85.5   \\
    policy gradient &  \bf 92.2 (+0.6)  & {\bf 83.3 (+0.6)} & {\bf 87.0 (+1.5)}  \\
    \bottomrule
  \end{tabularx}
  \caption{\label{table:alternative_comparisons}Test set F1 by training procedure, and in comparison to past work using the same models. Improvements over likelihood training are indicated in parentheses, with the highest results among the training procedures compared here in bold. $^*$: training uses a dynamic oracle; $^\dagger$: past work using a global scoring model (all models we train here are locally-normalized).
  }
\end{table}

\autoref{table:alternative_comparisons} compares parser F1 by training procedure for each language.
Policy gradient improves upon likelihood training in 14 out of 15 cases, with improvements of up to 1.5 F1. One of the three dynamic oracle-based training methods -- either likelihood with exploration, softmax margin (SMM), or softmax margin with exploration -- obtains better performance than policy gradient in 10 out of 12 cases. This is perhaps unsurprising given the strong supervision provided by the dynamic oracles and the credit assignment problem faced by policy gradient. However, a substantial fraction of this performance gain is recaptured by policy gradient in most cases.

While likelihood training with exploration using a dynamic oracle more directly addresses exploration bias, and softmax margin training more directly addresses loss mismatch, these two phenomena are still entangled, and the best dynamic oracle-based method to use varies. The effectiveness of the oracle method is also likely to be influenced by the nature of the dynamic oracle available for the parser. For example, the oracle for RNNG lacks F1 optimality guarantees, and softmax margin without exploration often underperforms likelihood for this parser. However, exploration improves softmax margin training across all parsers and conditions. 

Although results from likelihood training are mostly comparable between RNNG-128 and the larger model RNNG-256 across languages, policy gradient and likelihood training with exploration both typically yield larger improvements in the larger models, obtaining 92.6 F1 for English and 86.0 for Chinese (using likelihood training with exploration), although results are slightly higher for the policy gradient and dynamic oracle-based methods for the smaller model on French (including 83.5 with softmax margin with exploration). 
Finally, we observe that policy gradient also provides large improvements for the In-Order parser, where a dynamic oracle has not been defined.

We note that although some of these results (92.6 for English, 83.5 for French, 87.0 for Chinese) are state-of-the-art for single model, discriminative transition-based parsers, other work on constituency parsing achieves better performance through other methods. Techniques that combine multiple models or add semi-supervised data \cite{vinyals2015grammar,dyer2016rnng,choe16parsing,kuncoro2017syntax,liu2017inorder,fried2017improving} are orthogonal to, and could be combined with, the single-model, fixed training data methods we explore. Other recent work \cite{gaddy2018analysis,kitaev2018constituency} obtains comparable or stronger performance  with global chart decoders, where training uses loss augmentation provided by an oracle. By performing model-optimal global inference, these parsers likely avoid the exposure bias problem of the sequential transition-based parsers we investigate, at the cost of requiring a chart decoding procedure for inference.

Overall, we find that although optimizing for F1 in a model-agnostic fashion with policy gradient typically underperforms the model-aware expert supervision given by the dynamic oracle training methods, it provides a simple method for consistently improving upon static oracle likelihood training, at the expense of increased training costs.

\section*{Acknowledgments}

DF is supported by a Huawei / Berkeley AI fellowship.
This research used the Savio computational cluster provided by the Berkeley Research Computing program at the University of California, Berkeley.

\bibliography{refs}
\bibliographystyle{acl_natbib}

\clearpage
\appendix

\section{Supplemental Material}
\label{sec:supplemental}

\begin{table}[h]
  \centering

  \begin{tabular}{llll}
    \toprule
    training & English & French & Chinese \\
    \cmidrule(lr){1-1} \cmidrule(lr){2-4} 
    \multicolumn{4}{l}{Span-Based}  \\
    policy gradient & 10 & 10 & 10   \\
    likelihood+explore & 5 & 2 & 2  \\
    SMM+explore & 5 & 5 & 10 \\
    \hdashline \\[-0.8em]
    \multicolumn{4}{l}{Top-Down}  \\
    policy gradient & 5 & 2 & 10 \\
    likelihood+explore & 2 & 5 & 2  \\
    SMM+explore & 5 & 5 & 5 \\
    \hdashline \\[-0.8em]
    \multicolumn{4}{l}{RNNG Discriminative, Size 128 }  \\
    policy gradient & 10 & 10 & 10 \\
    likelihood+explore & 2 & 5 & 10 \\
    SMM+explore  & 2 & 2 & 2 \\
    \hdashline \\[-0.8em]
    \multicolumn{4}{l}{RNNG Discriminative, Size 256}  \\
    policy gradient & 10 & 10 & 10 \\
    likelihood+explore & 2 & 2 & 2 \\
    \hdashline \\[-0.8em]
    \multicolumn{4}{l}{In-Order}  \\
    policy gradient & 10 & 10 & 10   \\
    \bottomrule
  \end{tabular}
  \caption{\label{table:k_values} Number of candidates $k$ (including the gold tree for the sentence) used per sentence for each exploration technique.}
\end{table}

\end{document}